\documentclass{article}

\PassOptionsToPackage{numbers, compress}{natbib}
\usepackage[final]{nips_2018_1}

\usepackage[utf8]{inputenc} % allow utf-8 input
\usepackage[T1]{fontenc}    % use 8-bit T1 fonts
\usepackage{hyperref}       % hyperlinks
\usepackage{url}            % simple URL typesetting
\usepackage{booktabs}       % professional-quality tables
\usepackage{amsfonts}       % blackboard math symbols
\usepackage{nicefrac}       % compact symbols for 1/2, etc.
\usepackage{microtype}      % microtypography
\usepackage{graphicx}
\usepackage[inline]{enumitem}

\title{HYPE: A High Performing NLP System for Automatically Detecting Hypoglycemia Events from Electronic Health Record Notes}

\author{
  Yonghao Jin \\ 
  Department of Computer Science\\
  University of Massachusetts Lowell\\
  Lowell, MA 01854 \\
  \texttt{yonghao\_jin@student.uml.edu}\\
  \And
  Fei Li \\
  Department of Computer Science\\
  University of Massachusetts Lowell\\
  Lowell, MA 01854 \\
  \texttt{foxlf823@gmail.com} \\
  \AND
  Hong Yu \thanks{
    The Bedford Veterans Affairs Medical Center, Bedford, MA
    }\\
  Department of Computer Science\\
  University of Massachusetts Lowell\\
   Lowell, MA, 01854 \\
  \texttt{hong\_yu@uml.edu} \\
}

\begin{document}

\maketitle

\begin{abstract}
  Hypoglycemia is common and potentially dangerous among those treated for diabetes. Electronic health records (EHRs) are important resources for hypoglycemia surveillance. In this study, we report the development and evaluation of deep learning-based natural language processing systems to automatically detect hypoglycemia events from the EHR narratives. Experts in Public Health annotated 500 EHR notes from patients with diabetes. We used this annotated dataset to train and evaluate HYPE, supervised NLP systems for hypoglycemia detection. In our experiment, the convolutional neural network model yielded promising performance ( \(Precision=0.96 \pm 0.03, Recall=0.86 \pm 0.03, F1=0.91 \pm 0.03\) ) in a 10-fold cross-validation setting. Despite the annotated data is highly imbalanced, our CNN-based HYPE system still achieved a high performance for hypoglycemia detection. HYPE could be used for EHR-based hypoglycemia surveillance and to facilitate clinicians for timely treatment of high-risk patients.
\end{abstract}

\section{Introduction}

In 2014, Centers for Disease Control and Prevention (CDC) estimated that 29.1 million Americans aged 20 or older have diabetes mellitus (DM).\cite{CDC2011National2014} Treatment-associated hypoglycemia (low blood sugar) in patients with DM is the third most common adverse drug event (ADE), resulting in about 25,000 emergency department visits and 11,000 hospitalizations yearly among Medicare patients.\cite{lipskakjNAtionalTrendsUs2014} Electronic health records (EHRs) are important resources for reporting hypoglycemia.\cite{lipskakjNAtionalTrendsUs2014} However, studies have shown that many hypoglycemic events are not represented by ICD codes but are described in EHR narratives.\cite{DefiningReportingHypoglycemia2005a} Manual chart review is prohibitively expensive. Therefore, automatically extracting hypoglycemia-related information from EHR notes can be a crucial complement for extracting such information from structured EHR data.\cite{lipskakjNAtionalTrendsUs2014}

However, reliably detecting hypoglycemia events in EHR notes is very challenging for the following reasons. First, the descriptions of adverse events (in this case, hypoglycemia) can be very flexible in the clinical notes (e.g., “patient with hypoglycemia”, “she has low bs level”, “bs is in low 20”), so it is impossible to manually specify rules to recognize all the patterns. Second, hypoglycemia is a relatively rare adverse event, making it difficult to collect enough data to train a machine learning model.

In this paper, we report the development and evaluation of HYPE, natural language processing (NLP) systems to effectively detect hypoglycemia events in the EHR narratives. Our HYPE were built upon the advanced NN architectures, which have shown greatly outperformed traditional machine learning models in different NLP applications.\cite{rumengHybridNeuralNetwork2018, jagannathaStructuredPredictionModels2016, jagannathaBidirectionalRNNMedical2016} We have shown that our deep-learning-based HYPE systems were high-performing and state-of-the-art, outperforming the traditional machine learning models in a wide margin in both recall and precision. 

\section{EHR Corpus and Annotation}
With the approval from the Institutional Review Boards (IRB) at the University of Massachusetts Medical School, we annotated 500 English EHR notes from patients with diabetes who were treated at the UMass Memorial Center in 2015. Since hypoglycemia could be rare events\cite{lipskakjNAtionalTrendsUs2014, bellFrequencySevereHypoglycemia1997}, we therefore developed a strategy to increase the likelihood of mentioning in EHR notes. Specifically, we selected the first 500 notes by querying both hypoglycemia ICD-9 codes (251.*) and related diabetic medications (e.g., insulin and metformin). We split each note into sentences using the NLTK package\cite{birdNaturalLanguageProcessing2009}, which we have adapted to the EHRs. We asked experienced experts to annotate each sentence as containing a hypoglycemic event (Positive) or not (Negative). A sentence was annotated as Positive if it describes any hypoglycemia-related diagnosis and symptoms (e.g. “patients have low blood sugar level”). 

In all, the 500 EHR notes contains a total of 95,246 sentences (an average of \(190 \pm 114 \) sentences per note, minimum 6 and maximum 912) with 1,316 (3 \%) annotated as Positive sentences. The average sentence length is \(11.2 \pm 11\) (minimum 2 and maximum 318 ) words. We cropped sentences with more than 40 words for efficiency reason. All EHR notes were fully deidentified prior to annotation or being used to develop the NLP systems.

\section{Method}

\begin{figure}
  \centering
  \includegraphics[width=0.8\linewidth]{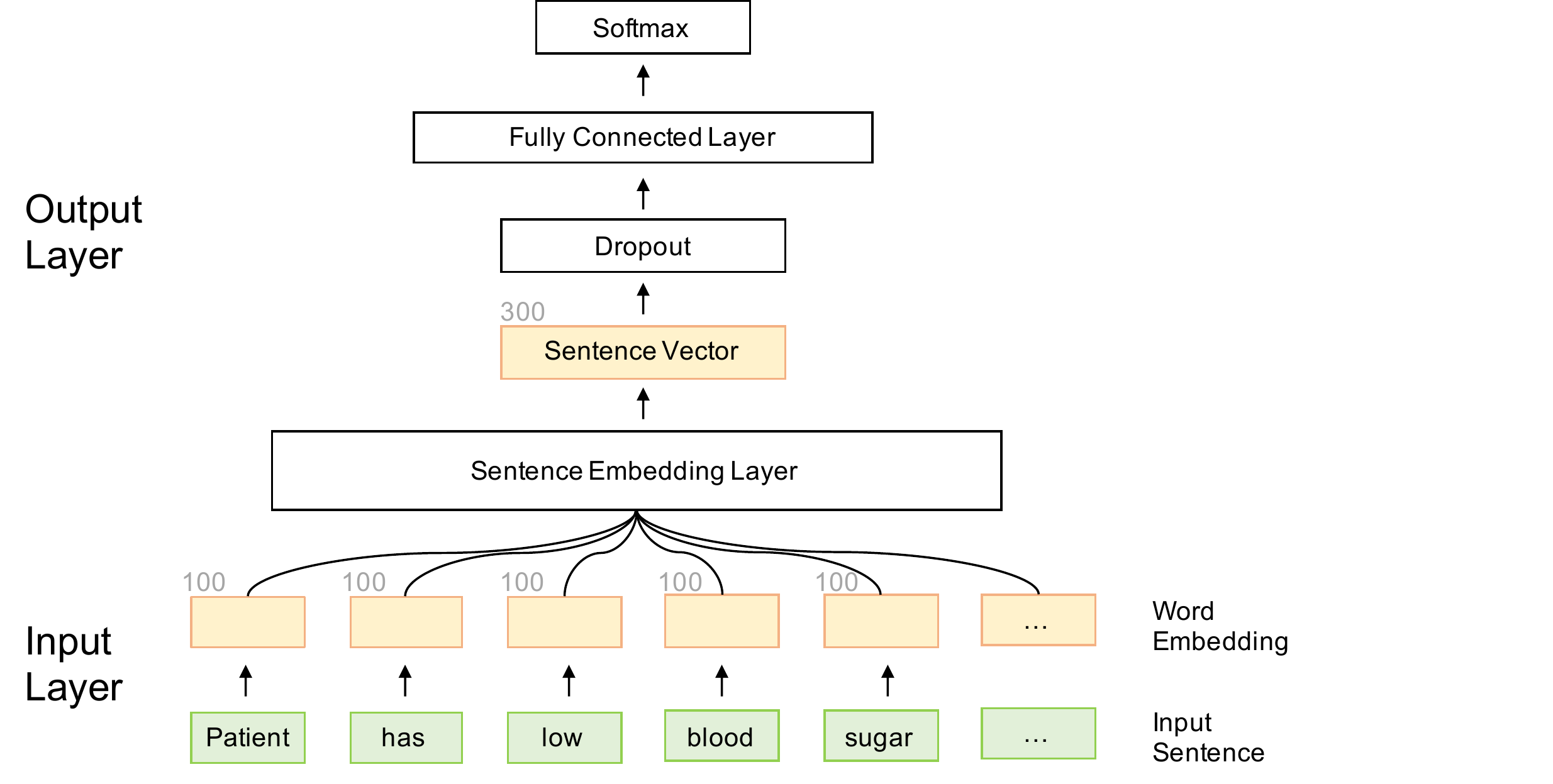}
  \caption{Model Architecture}
  \label{fig1}
\end{figure}

We used a standard deep learning text classification model\cite{kimConvolutionalNeuralNetworks2014}. The architecture of the model is shown in Figure~\ref{fig1}. The main architecture could be split into three parts: 
\begin{enumerate*} 
\item an input layer takes input sentence and constructs a matrix containing word embeddings for each word, 
\item a sentence embedding layer computes a fixed dimensional vector from the variable dimension matrix from last layer and 
\item the final output layer projects the sentence vector to probability scores of each class.
\end{enumerate*}

\begin{figure}
  \centering
  \includegraphics[width=1\linewidth]{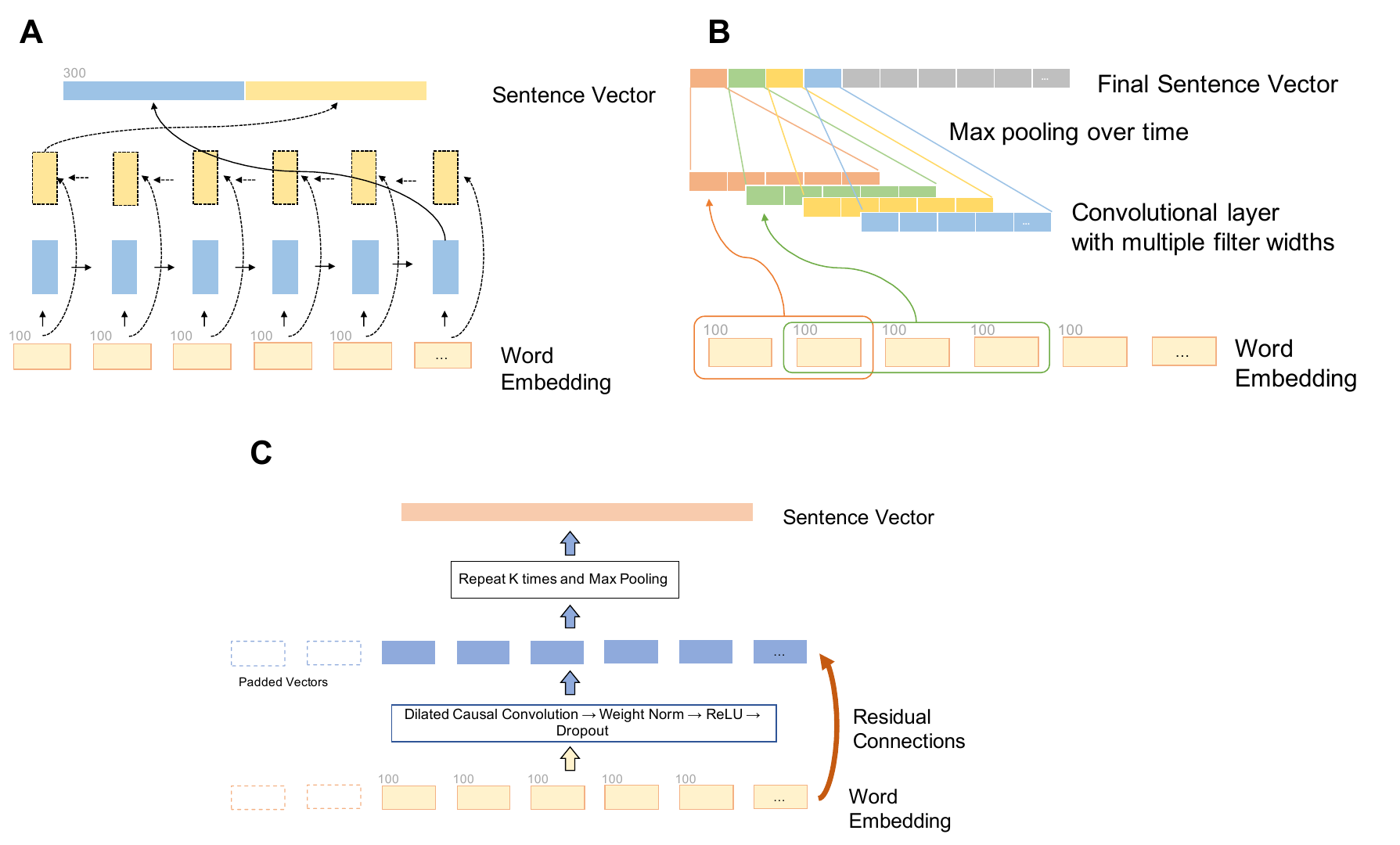}
  \caption{Sentence Embedding Layers. A. RNN Model B. CNN Model C. TCN Model}
  \label{fig2}
\end{figure}

We initialized the input layer with public available 100-dimension word vectors\cite{pyysaloDistributionalSemanticsResources2013} trained on a combined text corpus from both public and medical domains using Word2Vec. For the sentence embedding layer, we experimented with recurrent and convolutional neural network layer, shown in Figure~\ref{fig2}. In RNN setting, the output of the last step is chosen to be the sentence vector. In CNN, we applied max-pooling to each filter to build a fixed dimensional vector. The dimension of the sentence vector is fixed to be 300 in all experiments. In the output layer, the elements of the sentence vector is randomly zeroed out by a dropout layer with dropout rate 0.5 and finally, a softmax layer is used to build the probability model. During training, we used Adam algorithm\cite{kingmaAdamMethodStochastic2014} with learning rate 0.00005 on a cross-entropy loss. 

\section{Experiments}

\subsection{Evaluation}
Due to the sparsity of the dataset, we performed 10-fold cross-validation to robustly evaluate the performance of each model. The dataset was randomly split into 10 groups. Each time we took out one group as the testing set, and the rest as the training set. The development set was constructed by randomly separating 10\% data from each training set. 

\subsection{Baseline Model}
We applied SVM, a commonly used learning algorithm for classification problems, as our strong baseline model. Each sentence is vectorized by a long sparse vector with the dimension equal to the vocabulary size of the training corpus (after removing common stop words). We used the scikit-learn package to build the sentence vectors and train the SVM model with radial basis function kernel.\cite{pedregosaScikitlearnMachineLearning2011}

\section{Results and Discussion}
\begin{table}
  \centering
  \begin{tabular}{llllll}
    \toprule
    % \cmidrule(r){1-2}
    & SVM & LSTM     & BiLSTM & TCN & CNN \\
    \midrule
    Precision & \(0.74 \pm 0.07 \)  & \(0.91\pm0.02\) & \(0.91\pm0.02\) & \(0.92\pm0.03\) & $\mathbf{0.96\pm0.03}$   \\
    Recall     & \(0.57 \pm 0.05\) & \(0.86\pm0.02\) & \(0.87\pm0.04\) & $\mathbf{0.89\pm0.04}$ & \(0.86\pm0.03\)      \\
    F1     & \(0.64 \pm 0.03 \)  & \(0.88\pm0.02\) & \(0.88\pm0.02\) & \(0.90\pm0.02\) & $\mathbf{0.91\pm0.02}$  \\
    PR-AUC & $0.74$  & $0.93$ & $0.94$ & $0.96$ & $\mathbf{0.97}$\\
    ROC-AUC & $0.97$  & $0.996$ & $0.997$ & $\mathbf{0.998}$ & $\mathbf{0.998}$ \\
    \bottomrule
  \end{tabular}
  \caption{Detailed performance of each experiment}
  \label{table1}
\end{table}

\subsection{Principal Results}
Our results (Table~\ref{table1}) show that comparing with a strong-baseline SVM model, NN models all improved the performance (precision, recall, F1, PR-AUC, and ROC-AUC) by a large margin. The fundamental difference between a NN model and a SVM model is their representations of data. Our SVM models use bag-of-word and n-grams to represent the input sentences. In contrast, NN models are able to generate high-level sentence features, including both semantic and syntactic features. Our results also show that high-performance NN models can be trained using a relatively small set of annotated and imbalanced EHR data (a total of 41,034 sentences, of which 1,316 sentences are Positive instances). The implication is significant as the “knowledge-bottleneck” challenge has made it unrealistic to annotate a large amount of clinical data for supervised machine-learning applications.

\subsection{Comparison Between Different Neural Network Models}
Our results show that CNN performed the best for detecting sentence-level hypoglycemia, even though the data is imbalanced. One of the advantages of the recently proposed TCN model is its improved performance for longer-sentences. Therefore, it is not surprising that our results show that the TCN model yields the best recall \(0.89\pm0.04\), although its precision is lower than the CNN model. Our results show that the two RNN models (LSTM and Bi-LSTM), have similar performance and both under-performed the CNN models.  This suggests that RNN-based models are less effective than the CNN-models in capturing important sentence patterns for this task, even in a bi-directional configuration. The performance might improve by adding an attention mechanism, but that will greatly increase the complexities of the models. 

\subsection{Error Analysis}
We manually examined the error cases and identified two types of common errors. The models often failed on cases where hypoglycemia events are indicated in the numerical measurement of blood sugar level. While the models could easily identify sentences like “BS is low”, it almost always made mistakes when encountering “BS is 68” or “fsbs noted to be 9”. This type of sentences is difficult to be identified for many reasons. Frist, there is no good way to represent the relative size of a number in the embedding space, so it is impossible for the model to learn a “less than” operation to identify low blood sugar value. Second, the units of the numeric value are often absent, which must be inferred from the order of magnitude of the value. In the above examples, “68” should be “68 mg/dL” and “9” should be “9 nmol/L”. External human knowledge must be incorporated to correctly identify this kind of sentences. In the future, we may detect a unit expression and replace it with either “less than” or “more than.”

\subsection{Limitations and Future Work}
Our study has several limitations. Our annotated EHR data was small (a total of 500 EHR notes) and was selected using diabetes-related ICD codes, and therefore may not represent the natural distribution of the EHR data, in which the hypoglycemic events were sparse. To overcome this challenge, we focused on sentence-level classification, and therefore HYPE could be robust to the naturally distributed EHR data. On the other hand, because HYPE focused on sentence-level classification, it may miss hypoglycemic events that are expressed across multiple sentences. In future work, we may explore paragraph or document-level classification. Another limitation is that HYPE only detects the presence of a bleeding event. It does not identify assertion and severity. More annotated data is needed for refined classification and this will be our future work.

\section{Conclusion}
In this study, we addressed the question of how to automatically detect EHR note sentences containing hypoglycemia events. Our deep learning models also achieved both high precisions (up to about 96\%) and recalls (up to about 89\%). We found that among three deep learning models namely RNN, TCN, and CNN, CNN achieved the best performance. These encouraging results indicate that deep-learning-based approaches are promising to be applied to hypoglycemia event detection. Our work is an important step towards accurate surveillance of hypoglycemic events in EHRs and may help provide clinicians a valid tool to improve treatment of diabetes mellitus. 

\medskip

\small

\bibliographystyle{ieeetr}
\bibliography{hypoglycemia}{}

\end{document}